%% file: main.tex
\documentclass[conference]{IEEEtran}
\IEEEoverridecommandlockouts

\usepackage{package}
 
\def\BibTeX{{\rm B\kern-.05em{\sc i\kern-.025em b}\kern-.08em
    T\kern-.1667em\lower.7ex\hbox{E}\kern-.125emX}}

\begin{document}


\title{Large Language Models in the Travel Domain: An Industrial Experience
\thanks{This work has been partially supported by the Italian PNRR MUR project PE0000013-FAIR and PR CAMPANIA FESR 2021-2027 - Asse I - Obiettivo Specifico 1.1 – Azione 1.1.3 – Avviso pubblico CAMPANIA STARTUP 2023 (\texttt{CUP B68I23005640007}).}

}

\author{
\IEEEauthorblockN{
    Sergio Di Meglio \IEEEauthorrefmark{4},
    Aniello Somma \IEEEauthorrefmark{5},
    Luigi Libero Lucio Starace \IEEEauthorrefmark{4},\\
    Fabio Scippacercola \IEEEauthorrefmark{5},
    Giancarlo Sperlì \IEEEauthorrefmark{4},
    Sergio Di Martino \IEEEauthorrefmark{4}
    }

\IEEEauthorblockA{
    \IEEEauthorrefmark{4} Department of Electrical Engineering and Information Technology, University of Naples Federico II, Italy\\
    \IEEEauthorrefmark{5} Fervento srl, \\
    Email: (sergio.dimeglio, luigiliberolucio.starace, giancarlo.sperli, sergio.dimartino)@unina.it\\
    (aniello, fabio)@fervento.com
}}

\maketitle

\begin{abstract}
Online property booking platforms are widely used and rely heavily on consistent, up-to-date information about accommodation facilities, often sourced from third-party providers. However, these external data sources are frequently affected by incomplete or inconsistent details, which can frustrate users and result in a loss of market.

In response to these challenges, we present an industrial case study involving the integration of Large Language Models (LLMs) into \textsc{Caleidohotels}, a property reservation platform developed by \textsc{Fervento}. We evaluate two well-know LLMs in this context: Mistral 7B, fine-tuned with QLoRA, and Mixtral 8x7B, utilized with a refined system prompt. Both models were assessed based on their ability to generate consistent and homogeneous descriptions while minimizing hallucinations.

Mixtral 8x7B outperformed Mistral 7B in terms of completeness (99.6\% vs. 93\%), precision (98.8\% vs. 96\%), and hallucination rate (1.2\% vs. 4\%), producing shorter yet more concise content (249 vs. 277 words on average). However, this came at a significantly higher computational cost: 50GB VRAM and \$1.61/hour versus 5GB and \$0.16/hour for Mistral 7B.
Our findings provide practical insights into the trade-offs between model quality and resource efficiency, offering guidance for deploying LLMs in production environments and demonstrating their effectiveness in enhancing the consistency and reliability of accommodation data.


\end{abstract}


\input{Sections/introduction}
\input{Sections/background}
\input{Sections/motivation}

\input{Sections/framework}
\input{Sections/evaluation}

\input{Sections/related}
\input{Sections/conclusions}

\balance
\bibliographystyle{IEEEtran}
\bibliography{bibliography}

\end{document}

%% file: Sections/introduction.tex
\section{Introduction}
The travel industry is a highly competitive and rapidly evolving sector in which maintaining accurate, up-to-date, and complete data is critical to success. Online travel platforms must present information from multiple sources, often with varying degrees of completeness and quality \cite{doi:10.1177/1096348020980101}. \textsc{caleidohotels}\footnote{https://caleidohotels.com/welcome}, an online property booking platform, faces the ongoing challenge of aggregating facility data from multiple providers, ensuring that users receive a consistent set of information. This consistency is essential not only to improve the user experience but also to maintain the platform's competitive differentiation in the marketplace \cite{di2025performance,di2024evaluating}.

As the capabilities of LLMs have advanced, interest has grown in their application in various industries, including travel \cite{tourbert}. LLMs offer a promising solution to address the challenges of data aggregation and content generation by leveraging their ability to process and generate human-like text \cite{vito2024large}. However, implementing LLMs in a dynamic and specialized domain such as travel requires careful consideration of the models to be used, the data to be processed, and the specific tasks to be performed \cite{starace2024can}.

In this paper, we explore the application of LLMs to improve the quality and consistency of facility descriptions on \textsc{caleidohotels}. To this end, we considered two distinct strategies: (1) the fine-tuning of a smaller, resource-efficient model (Mistral 7B) using the QLoRA \cite{dettmers2023qlora} technique, and (2) the use of a larger, more powerful model (Mixtral 8x7B) through a refined system prompt. The goal was to create a robust system capable of generating high-quality property descriptions, filling gaps in the existing catalog, and ensuring uniform presentation of properties across the platform. The main contributions of this are:

\begin{enumerate}[leftmargin=*]
    \item We describe our industrial experience, discussing the practical steps we put in place to integrate and assess LLMs into an industrial booking platform, and providing a detailed roadmap for similar implementations in the travel domain or different industrial domains.
    \item We present an empirical evaluation involving two state-of-the-art LLMs with different characteristics: a smaller but fine-tuned model (Mistral 7B), and a larger model (Mixtral 8x7B). We discuss the fine-tuning and prompt engineering processes, and the trade-offs of each approach, providing useful insights to researchers and practitioners interested in similar applications.
    \item We show that LLMs have the potential to be an important asset in the considered industrial scenario, improving the consistency and reliability of facility descriptions, and thereby enhancing the overall user experience. 
\end{enumerate}

%% file: Sections/motivation.tex
\section{Industrial Motivations and Objectives}
\label{sec:industrial_motivations}

\textsc{Fervento}\footnote{\url{https://fervento.com}} is a startup developing software products across various domains, including travel. In 2023, it launched \textsc{Caleidohotels}, a global hotel and property booking platform allowing users to browse hundreds of listings. A standout feature is its proprietary engine, which, supported by an interactive map, aggregates accommodation offerings from multiple providers, enabling users to find the best rates for services they need.

Given the dynamic and fragmented nature of the travel industry \cite{doi:10.1177/1096348020980101}, ensuring a consistent and high-quality user experience is critical \cite{di2023starting, e2e2025}. This requires integrating heterogeneous data sources into a unified interface to enhance readability and comparison.
However, catalogs from multiple providers often lack uniformity in how facilities are described. Properties with sparse or inconsistent descriptions are frequently overlooked by users, resulting in lost opportunities \cite{vila2021indicators}. To address this, platforms must offer clear, coherent, and comparable facility descriptions.
Two possible solutions include incorporating third-party catalogs or developing a proprietary one. While external catalogs may be outdated or lack distinctiveness, building an in-house catalog is resource-intensive due to the need for frequent updates.

Previously, \textsc{Fervento} prioritized a main provider’s catalog and used secondary sources only when necessary. However, fallback descriptions often lacked the quality of the primary source, negatively impacting the user experience.
With the rise of LLMs, \textsc{Fervento} explored their application to enhance accommodation descriptions. This paper presents CaleidoGen, a system designed to overcome the limitations described above. By leveraging LLMs, in particular, Generative AI (GenAI), it offers several advantages:

\begin{itemize}[leftmargin=*]
\item \textbf{Enhanced User Experience}: Unified descriptions across providers present information consistently, giving a cohesive brand feel and simplifying comparison between options. 
\item \textbf{Customization and Personalization}: Descriptions can highlight specific features, such as unique amenities or proximity to local attractions, tailored to seasonal interests (e.g., swimming pools or beaches for summer). 
\item \textbf{Ease of Extension}: A dedicated generation module simplifies the integration of new or lower-quality data sources, enabling expansion with minimal overhead. 
\item \textbf{Differentiation from Competitors}: A proprietary, easily updated content engine allows \textsc{Caleidohotels} to stand out with unique descriptions, turning the catalog into a strategic asset. 
\end{itemize}

In summary, the system not only enhances the likelihood of customer engagement and purchase but also positions \textsc{Caleidohotels} as a distinctive and competitive player in the online booking market.

%% file: Sections/framework.tex
\section{Experience of CaleidoGen}\label{sec:caleidogen}
As mentioned above, \textsc{Fervento} decided to use GenAI to efficiently create a catalogue of coherent structures from existing data sources, with the aim of reducing resource utilisation compared to traditional methods. However, exploiting GenAI effectively requires specialized skills and is not as simple as using standard tools.

To achieve practical knowledge in this area, \textsc{Fervento} decided to follow an iterative process, focusing on different aspects of the data transformation to gradually derive the system architecture. Each of these aspects has been evaluated by small proofs-of-concept that analyzed the (sub-)problems, including implementation and testing activities.

\begin{figure}[h]
    \centering
    \includegraphics[width=\columnwidth]{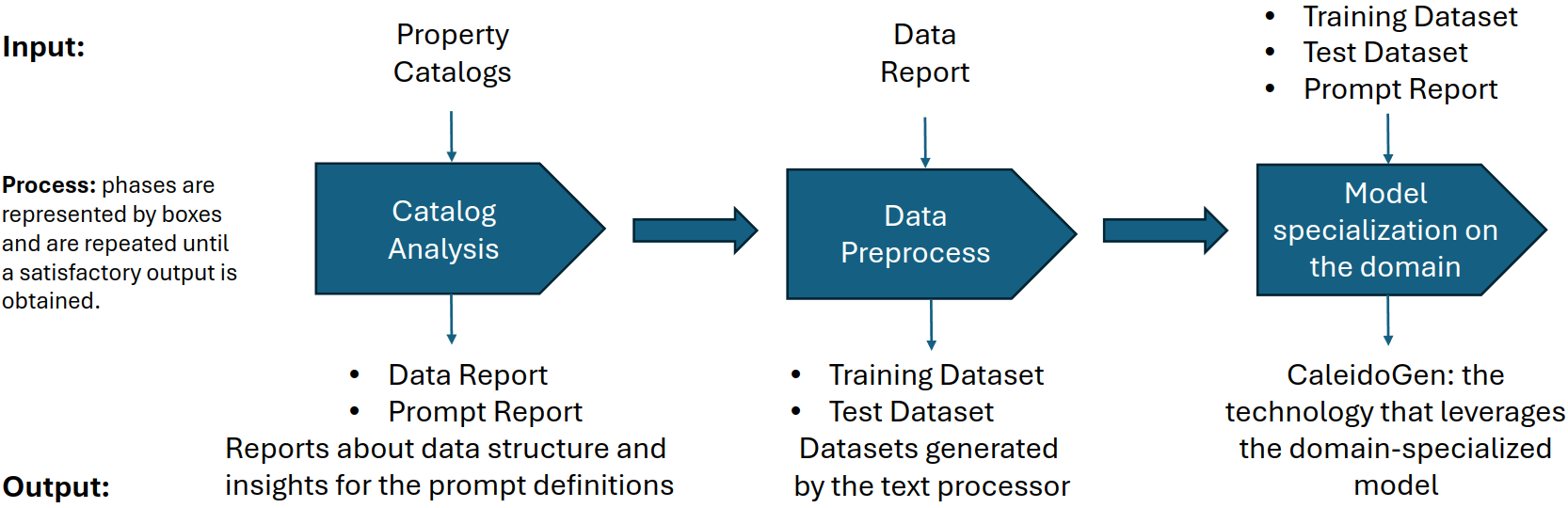}
    \caption{CaleidoGen's iterative process}
    \label{fig:pipeline}
\end{figure}

This iterative process, aimed at producing such technology, is characterized by 3 phases, as shown in Figure \ref{fig:pipeline}.
Phase 1 aimed to analyze the available input data sources, which consisted of 3 catalogs. 
At this stage, it was necessary to manually explore the quality of the catalogs and their differences to learn their completeness, the difference in their writing style, and the ability of parsers to extract or reshape their content efficiently. The output of this phase consisted of a Prompt Report, a design feedback about how to write an effective GenAI prompt (for Phase 3), and a Data Report, a design feedback regarding the data model of the catalogs facility, useful to assess the output of CaleidoGen (for Phase 2). 

Phase 2 focused on investigating how to adapt the raw text catalog in the best shape to be processed by a GenAI and the benefits of the model's fine-tuning. In this stage, the training and the test datasets were developed thanks to the text processor, which is responsible for properly preprocessing the data. The production of these datasets allowed the Phase 3 work to be defined.  

Phase 3 focused on tuning the best domain-specialized model to enhance the quality of the facility description output. Once the input data were available, multiple implementations were tested, including LLM models, parameters, and their runs. Finally, the best results achieved in each stage resulted in the final CaleidoGen architecture, which yielded a notable enhancement in the functionality of \textsc{caleidohotels}.

\subsection{Data Preprocessing}
\label{data-preprocess}
One of the output of the catalog analysis phase was the input for Phase 2, the \textit{Data Preprocessing} phase. The latter serves two primary purposes: (i) to create a training dataset for the model's fine-tuning, and (ii) to extract the necessary context for creating the test dataset. Developing a dedicated dataset for further training the model is crucial \cite{huang2015empirical, datapreprocess}, as it directly influences the model's ability to understand and learn, particularly in the context of content generation. Poor preprocessing can lead to the model learning incorrect information and exhibiting hallucinations, resulting in suboptimal performance \cite{huang2023survey, di2024visual}.

The training dataset, consisting of 100 items and needed for fine-tuning, was generated from a subset of data provided by the primary catalog, after a cleaning phase. This step was particularly tricky due to the presence of irregular formatting, HTML tags, and inconsistent units of measure. These flaws could have compromised the effectiveness of the fine-tuning process. With the feedback from Phase 1, it was possible to perform this data cleansing efficiently. 

The next step focused on extracting the features of the facilities listed in the catalog. This allowed us to generate the "\textit{Context}" field in our dataset, which provides the model with the necessary characteristics of each property to accurately generate the corresponding descriptions. The context is constructed by organizing these features into detailed categories, including Recreation, Services, Dining, Rooms, Additional Services, and nearby Points of Interest (POIs). This structured approach ensures that the model has a comprehensive understanding of each property's offerings, enabling it to produce more accurate and engaging descriptions.



\begin{figure}[h]
    \centering
    \includegraphics[width=0.8\columnwidth]{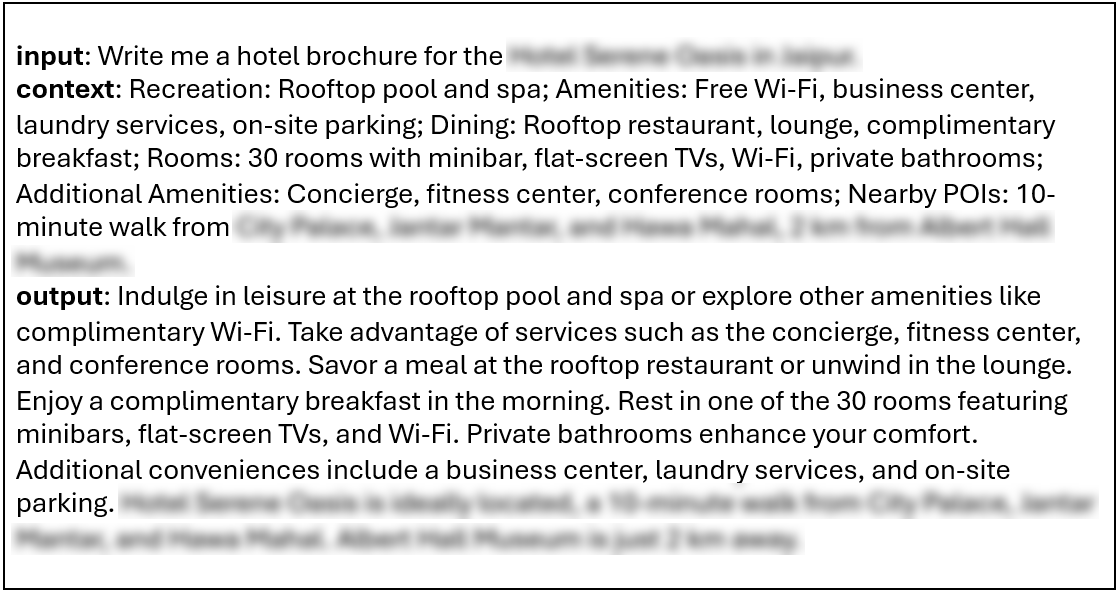}
    \caption{This example presents the formatting of a dataset object}
    \label{fig:example-dataset}
\end{figure}

The dataset objects are structured as follows:

\begin{itemize}
    \item \textbf{Input}: Represents the request for the facility description.
    \item \textbf{Context}: A set of features and services offered by the specific facility.
    \item \textbf{Output}: The final facility description obtained from the catalog after the processing step.
\end{itemize}

In our prototype runs, this structuring of the context field proved to be the best. This structured approach ensures that the model is trained on relevant and consistent data, an example is shown in Figure \ref{fig:example-dataset}. 


\subsection{Exploration of CaleidoGen's LLMs
}
One of the main focuses of the activity was to select a model that provided a good compromise between performance and hardware requirements to obtain a useful evaluation for industrial exploitation. To select such a model, it was decided to test the performance of two different models.

Initially, Mistral 7B, the smallest model in the Mistral family, was chosen. Then, a larger and more complex model was selected, Mixtral 8x7B, which is distinguished by its different architecture and significantly more parameters. The models are both open source and are known for their use in various industries \cite{moslem2023fine}, their effectiveness and reliability were also confirmed in the literature \cite{ono2024evaluating}. Mistral 7B was configured with a 4-bit quantization on which the fine-tuning was performed. Successively, a comparison was made with the results obtained with Mixtral 8x7B, configured with 8-bit quantization, on which the system prompt was refined. 



The following subsections give more details about the two models, particularly their prompt definition process, and in addition for Mistral 7B also the fine-tuning process.

\subsubsection{Mistral 7B with QLoRA Fine-Tuning}

Mistral 7B is a 7-billion parameters language model that has outperformed many other open-source models \cite{open-llm-leaderboard-v2}. It uses innovative techniques, such as grouped-query attention and sliding window attention, to improve inference speed and reduce memory requirements \cite{jiang2023mistral7b}, striking a good balance between overall performance and hardware requirements. 

The prototype runs of Mistral 7B were computed on an Amazon Web Services (AWS) virtual machine instance equipped with the NVIDIA Tesla T4 GPU with 16 GB of VRAM. The choice to load the model with the 4-bit quantized configuration was necessary to take full advantage of this resource during model loading and inference. 

%
%
%
The fine-tuning technique used on the Mistral 7B model was QLoRA, which proved to be one of the most used in the literature \cite{LLMSurvey}. The latter is an LLM optimization technique that relies on loading a quantized version of the model. This quantization takes the form of using different data types in the loading of weights, thus decreasing accuracy but without sacrificing performance \cite{dettmers2023qlora}. The execution of a single prototype run took about four hours. In the remainder of this document, we will refer to the fine-tuned model of Mistral 7B as Mistral 7B-FT.

The next section describes the steps for the prompt definition of this model.

\paragraph{Prompt Definition}
 
\begin{figure}[htbp]
    \centering
    \includegraphics[width=0.65\columnwidth]{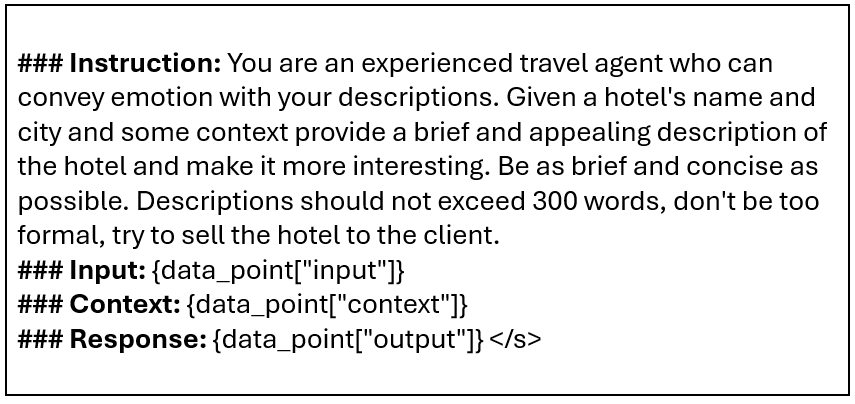}
    \caption{Example of the finetuning prompt of Mistral 7B}
    \label{fig:prompt-mistral7b}
\end{figure}

In language model optimization, the prompt is not only a means of communication with the model but is a key element that directs the model's behavior toward the desired goal \cite{googleprompteng}. Its structure and content can significantly influence the accuracy and relevance of the generated responses, as seen from the different prototype runs performed. With this knowledge, and with the prompt insights derived from the Prompt Report of Phase~1, our prompt was defined to reflect the structure of the dataset used during fine-tuning, as shown in figure \ref{fig:prompt-mistral7b}. The “context” field of the prompt is constructed using the same process described in Section \ref{data-preprocess}, but with a generalized approach to make it suitable to all the providers, considering that they offer data in different formats and structures, as shown in the Data Report of Phase 1.

\subsubsection{Mixtral 8x7B}

Mixtral 8x7B is a sparse mixture of expert models (SMoE) with open weights, licensed under the Apache 2.0 open-source software license. It has outperformed models like Llama 2 70B and GPT-3.5 in most benchmark tests \cite{jiang2024mixtralexperts}, making it a strong candidate for applications requiring high performance without proprietary constraints. 


The prototype runs of Mixtral 8x7B were computed on an Amazon Web Services (AWS) virtual machine instance equipped with four NVIDIA L4 GPUs, each with 24 GB of VRAM, for a combined total of 96 GB. The utilization of these GPUs in parallel permitted the model to be loaded in a quantized version, specifically with the 8-bit configuration. 

To utilize parallelization on multiple GPUs in an optimal manner, it was necessary to manually define a \textit{device map}, which indicates which layers should be loaded on which device among those available. This issue arises because the standard allocation provided by several libraries (in particular, the \textit{accelerate} library \cite{accelerate}) tends to distribute the model equally across the various available GPUs. However, this does not necessarily guarantee an optimal distribution of the latter because certain layers can saturate the memory of the device to which they are assigned at inference time. This identified issue, given also the shortcomings of the state of the art, required a thorough investigation to be able to identify the final solution. This was necessary since these technologies and techniques are still in a very early stage of development.



\begin{figure}
    \centering
    \includegraphics[width=0.8\columnwidth]{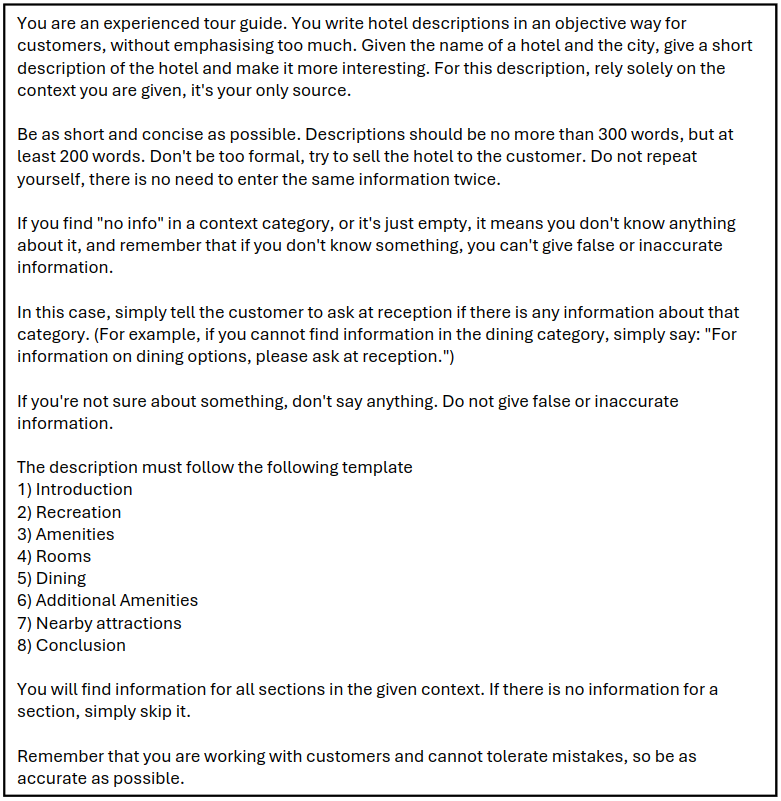}
    \caption{Example of System prompt used with Mixtral 8x7B}
    \label{fig:sysprompt}
\end{figure}

\paragraph{System Prompt Definition}
Using a model with a significantly larger number of parameters, such as Mixtral-8x7B, allowed us to adopt a completely different approach compared to the previous model. As the number of parameters increases, so does the model's capacity to handle more complex tasks and generate more nuanced output. This enhanced capability made it feasible to fully leverage the model's strengths through the refinement of the System Prompt. The latter is a special instruction within the conversation, designed to guide the model's behavior by providing explicit directives. By refining the System Prompt, as illustrated in Figure \ref{fig:sysprompt}, we were able to tailor the model's responses to better suit the specific tasks required.
The structure of the prompt for Mixtral-8x7B retained the same “input” and “context” fields as the prompt used for Mistral 7B, with the main difference being the exclusion of a dedicated “instruction” field and the adoption of a system prompt. A crucial step was modifying the tokenizer, which converts text into tokens the model can process \cite{hf-tokenizer}. 

The original Mixtral tokenizer did not support the “SYSTEM” role in messages due to limitations in its Jinja-based template. To address this, it was necessary to customize the tokenizer template to properly support and handle the system prompt, ensuring accurate communication with the model. This analysis was necessary to address this limitation of the state of the art, again due to the early state of these technologies.

%% file: Sections/evaluation.tex
\section{Empirical Evaluation}\label{sec:evaluation-design}

The evaluation aimed to compare the performance of the Mistral 7B-FT and Mixtral 8x7B models in generating high-quality facility descriptions. This comparative analysis was designed to assess the trade-offs between computational resource requirements and the quality of the generated content, ultimately identifying the most suitable model to be integrated into the \textsc{caleidohotels} system.
To conduct the evaluation, 20 facility descriptions were generated using each model. These facilities were not included in the optimization dataset but were part of a distinct testing dataset, ensuring an unbiased assessment of the model's ability to generalize to new data. 



\begin{figure}[htbp]
    \centering
    \includegraphics[width=0.8\linewidth]{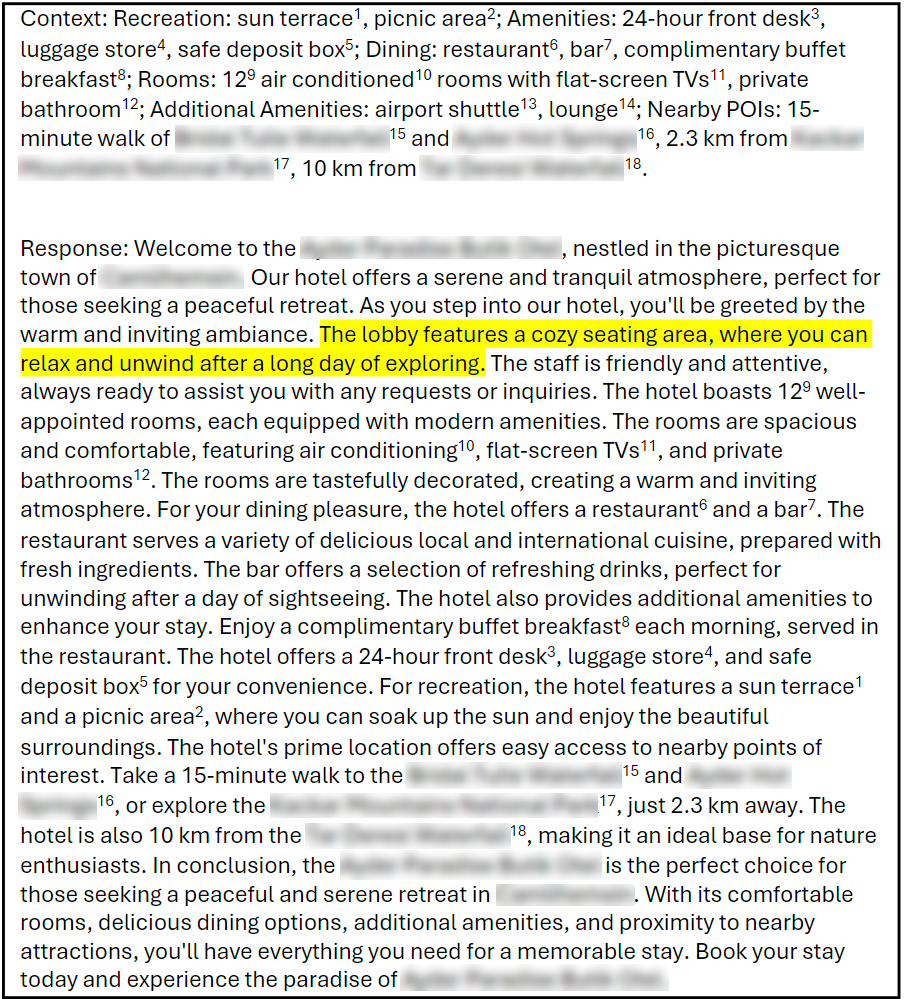}
    \caption{Analysis of the generation}
    \label{fig:genanalysis}
\end{figure}

To evaluate the generated descriptions, two of the authors manually mapped each feature present in the context to the corresponding elements in the generated description, carefully tracking any instances of hallucinations. Figure \ref{fig:genanalysis} shows a detailed analysis of a facility description, highlighting in yellow the segments identified as hallucinations.

The process was repeated five times to account for fluctuations due to the non-deterministic behavior of LLMs \cite{10.1145/3643661.3643952}. The metrics described below were then calculated for each repetition.


\subsection{Metrics}
Based on the analysis of the generated descriptions, we employed several metrics to numerically evaluate the quality of the output. These metrics, utilized in other studies within the literature \cite{song2024finesurefinegrainedsummarizationevaluation, hu2024unveilingllmevaluationfocused}, were carefully chosen to measure the performance of the models, offering a systematic and objective assessment of the generated content. 
The following evaluation metrics were introduced:
\begin{itemize}[leftmargin=*]
    \item \textbf{Completeness}: The initial evaluation metric pertains to the completeness of the text, that is, the extent to which context information is incorporated within the description
    \[ \text{\textit{Completeness}} = \frac{\text{\textit{Context Features Added}}}{\text{\textit{Total Context Features}}} \times 100 \]
    
    \item \textbf{Precision}: This evaluation metric focuses on the factual integrity of the text, assessing the extent to which the information presented in the description aligns with reality.
    \[ \text{\textit{Precision}} = \frac{\text{\textit{Correct Features Added}}}{\text{\textit{Total Features Added}}} \times 100 \]
    
    \item \textbf{Hallucinations}: This evaluation metric focuses on the presence of hallucinations within the description generated.
    \begin{equation*}
    \text{\textit{Hallucinations}} = \frac{\text{\textit{Hallucinated Features}}}{\text{\textit{Total Features Added}}} \times 100
    \end{equation*}
    \item \textbf{Length of generation}: The final metric under consideration is the length of generation in terms of the number of tokens, which in this context refers to the number of words in the text.
\end{itemize}

\subsection{Results}
The results of our study are summarized in \autoref{tab:results}. The table reports, for each model and the metrics, the overall average and standard deviation calculated across the twenty considered accommodation structures.
The results demonstrate the significant potential of LLMs in improving the quality and consistency of property descriptions on an online hotel booking platform such as \textsc{caleidohotels}. Both Mistral 7B-FT and Mixtral 8x7B models showed excellent results, making them good candidates for integration into the platform. However, to decide which model to integrate, we had to conduct a more thorough evaluation to find a trade-off.

From the perspective of content generation, as shown in Table~\ref{tab:results}, both models produced high-quality descriptions, but several distinctions were apparent. Mixtral 8x7B consistently outperformed Mistral 7B-FT across nearly all metrics. In terms of completeness, both models performed well, with Mixtral 8x7B achieving an impressive 99.6\% compared to 93\% for Mistral 7B-FT, reflecting its superior ability to capture relevant features in the descriptions. 

When considering precision, Mixtral 8x7B again led with 98.8\%, surpassing Mistral 7B-FT’s 96\%, showcasing its higher accuracy in representing real-world data and reducing errors. The hallucination metric further highlighted Mixtral 8x7B’s advantage, with a much lower rate of fabricated details at just 1.2\%, compared to Mistral 7B-FT’s 4\%, enhancing the reliability of the content generated. Although Mixtral 8x7B produced slightly shorter descriptions, averaging 249.2 words compared to 277 for Mistral 7B-FT, this did not compromise the inclusion of essential details. Instead, it demonstrated the model's ability to generate more concise and readable content without sacrificing quality.

From a computational standpoint, Mixtral 8x7B proved to be significantly more resource-intensive, requiring approximately 50GB of VRAM, roughly ten times more than the 5GB needed for Mistral 7B-FT. Running Mixtral 8x7B also necessitated a higher number of GPUs to operate effectively. While this increase in resource demand translated into enhanced output quality, making it a strong candidate for high-performance production environments, it also posed challenges in terms of infrastructure and cost.

On the other hand, Mistral 7B-FT was much less resource-hungry, offering a more cost-effective solution. Requiring only 5GB of VRAM, it remained competitive in terms of performance, particularly when fine-tuned using QLoRA. This makes Mistral 7B-FT a viable option in scenarios where computational resources are more constrained, providing a balanced approach between quality and operational efficiency. The AWS instances utilized for Mistral 7B-FT (1) and Mixtral 8x7B (2) shared the following common characteristics: they were equipped with the Linux operating system and were utilized on a spot basis. The hourly cost of instance (1) is \$$0.1606$, while that of instance (2) is \$$1.6121$.


\begin{table}
  \centering
    \scriptsize  
    \setlength{\aboverulesep}{0pt}
    \setlength{\belowrulesep}{0pt}
    \setlength{\extrarowheight}{.2ex}
    \arrayrulecolor{black}
    \caption{Average and Standard Deviation of Evaluation Metrics for Mistral 7B-FT and Mixtral 8x7B}
    \label{tab:results}
    \begin{tabular}{lcccc}
    \toprule
    \rowcolor{arsenic}
    \textcolor{white}{\textbf{Model}} & \textcolor{white}{\textbf{Completeness}} & \textcolor{white}{ \textbf{Precision}} & \textcolor{white} {\textbf{Length}} & \textcolor{white}{\textbf{Hallucinations}} \\
    \midrule
    \rowcolor{gray!5}
    \makecell{Mistral\\7B-FT}   & (93\% – 8.8\%)     & (96\% – 3.2\%)     & (277 – 70)      & (4\% – 3.8\%)     \\
    \midrule
    \rowcolor{white}
    \makecell{Mixtral\\ 8x7B}   & (99.6\% – 1.4\%)   & (98.8\% – 3.2\%)   & (249.2 – 28)    & (1.2\% – 3.2\%)   \\
    \bottomrule
  \end{tabular}
\end{table}

%% file: Sections/related.tex
\section{Related Work}\label{sec:related}

The emergence of ChatGPT and other LLMs has generated considerable interest in their applications in the tourism industry. 
\cite{wei2024tourllm} presents TourLLM, a fine-tuned model addressing the limitations of general-purpose LLMs in tourism contexts, especially in Chinese. They introduce the Cultour dataset, composed of tourism knowledge base QA pairs, travelogues, and diverse tourism-related QAs. \cite{HSU2024103723} explores opportunities and challenges of GenAI in enhancing tourism experiences and operations, proposing a conceptual framework and stressing the importance of fine-tuning and ethical considerations. \cite{doi:10.1080/19368623.2023.2211993} examines how ChatGPT can transform customer interactions and decision-making in tourism, while also discussing risks like job displacement and disruptions to traditional service models. \cite{doi:10.1080/02508281.2023.2287799} offers a theoretical framework outlining key propositions for ChatGPT adoption, focusing on customer experience, engagement, and trust.
Despite the growing interest, the number of articles exploring the topic remains limited. Few studies have developed models that can be directly integrated into existing systems, with most focusing on the wider market implications, including the travel industry.

%% file: Sections/conclusions.tex
\section{Conclusions}\label{sec:conclusions}

In conclusion, integrating LLM-based technologies into \textsc{Caleidohotels} effectively addressed the challenge of maintaining a consistent and complete catalog of property descriptions, particularly for providers lacking detailed content. 
The implementation of CaleidoGen enabled the generation of uniform, enriched descriptions that include facility details, room features, and nearby points of interest, enhancing both content quality and user experience, while helping the platform stand out in the competitive online booking market. 
The evaluation of Mistral 7B-FT and Mixtral 8x7B showed that both models are suitable for integration, with Mixtral 8x7B exhibiting superior performance in terms of completeness, accuracy, and reduced hallucinations. 
Hence, despite its higher computational costs, Mixtral 8x7B was adopted as the core model for \textsc{Caleidohotels}, also due to the importance of providing accurate information in this domain. 

These findings highlight the value of advanced LLMs in solving complex challenges in tourism and technology. They also serve as a broader call to action for companies in other industries to explore similar applications. This work marks only the beginning, with future developments including LLM-driven features such as generating guided tour plans based on nearby attractions or creating personalized itineraries from user preferences.